\documentclass[10pt,twocolumn,letterpaper]{article}

\usepackage{iccv}
\usepackage{times}
\usepackage{epsfig}
\usepackage{graphicx}
\usepackage{amsmath}
\usepackage{amssymb}
\usepackage{cite}
\usepackage{algpseudocode}
\usepackage{amsmath,amssymb,amsfonts}
\usepackage{graphicx}
\usepackage{textcomp}
\usepackage{xcolor}
\usepackage[ruled,vlined]{algorithm2e}
\usepackage{multirow}
\usepackage{multirow}
\usepackage{authblk}
\usepackage{subfig}

\usepackage[font=small,labelfont=bf]{caption}

\DeclareCaptionFont{small}{\small}

\usepackage[breaklinks=true,bookmarks=false]{hyperref}

\iccvfinalcopy % *** Uncomment this line for the final submission

\def\iccvPaperID{TradiCV-9} % *** Enter the ICCV Paper ID here

\ificcvfinal\fi

\begin{document}

%%%%%%%%% TITLE
\title{CoverTheFace: face covering monitoring and demonstrating using deep learning and statistical shape analysis}

% \author{Pengyue Hou \\
% University of Alberta\\
% {\tt\small pengyue@ualberta.ca}
% \and
% Ming(Chloe) Zhou\\
% University of Alberta\\
% {\tt\small mzhou4@ualberta.ca}
% \and
% Xingyu Li\\
% University of Alberta\\
% {\tt\small xingyu@ualberta.ca}
% \and
% Han Jie\\
% University of Alberta\\
% {\tt\small jhan8@ualberta.ca}
% \and
% Petr Musilek\\
% University of Alberta\\
% {\\tt\small pmusilek@ualberta.ca}
% }
\author[ ]{Yixin Hu}
\author[ ]{Xingyu Li}
\affil[ ]{Department of Electrical and Computer Engineering, University of Alberta}
\affil[ ]{\textit {\{yhu17,xingyu\}@ualberta.ca}}
\setcounter{Maxaffil}{0}
\renewcommand\Affilfont{\itshape\small}

\maketitle
\ificcvfinal\thispagestyle{empty}\fi

%%%%%%%%% ABSTRACT
\begin{abstract}
Wearing a mask is a strong protection against the COVID-19 pandemic, even though the vaccine has been successfully developed and is widely available. However, many people wear them incorrectly. This observation prompts us to devise an automated approach to monitor the condition of people wearing masks. Unlike previous studies, our work goes beyond mask detection; it focuses on generating a personalized demonstration on proper mask-wearing, which helps people use masks better through visual demonstration rather than text explanation. The pipeline starts from the detection of face covering. For images where faces are improperly covered, our mask overlay module incorporates statistical shape analysis (SSA) and dense landmark alignment to approximate the geometry of a face and generates corresponding face-covering examples. Our results show that the proposed system successfully identifies images with faces covered properly. Our ablation study on mask overlay suggests that the SSA model helps to address variations in face shapes, orientations, and scales. The final face-covering examples, especially half profile face images, surpass previous arts by a noticeable margin.
\end{abstract}

%%%%%%%%% BODY TEXT
\section{Introduction}
Masking face correctly reduces the spray of respiratory droplets. Particularly, it has been proven to be an effective measure of protection against the COVID-19 epidemic. According to the World Health Organization (WHO) guidelines, correctly masking is defined as all nose, mouth, and chin are covered. However, many people refuse to wear masks or wear them incorrectly, for example, wearing masks without covering their noses. According to an infection control epidemiologist with the University of Toronto, it is the same as not wearing masks if wearing them incorrectly. Therefore, checking people if they are wearing masks correctly in public, densely populated spaces has become a significant problem.

\captionsetup[figure]{font=small}
\begin{figure}\smaller
\centerline{\includegraphics[width=\linewidth]{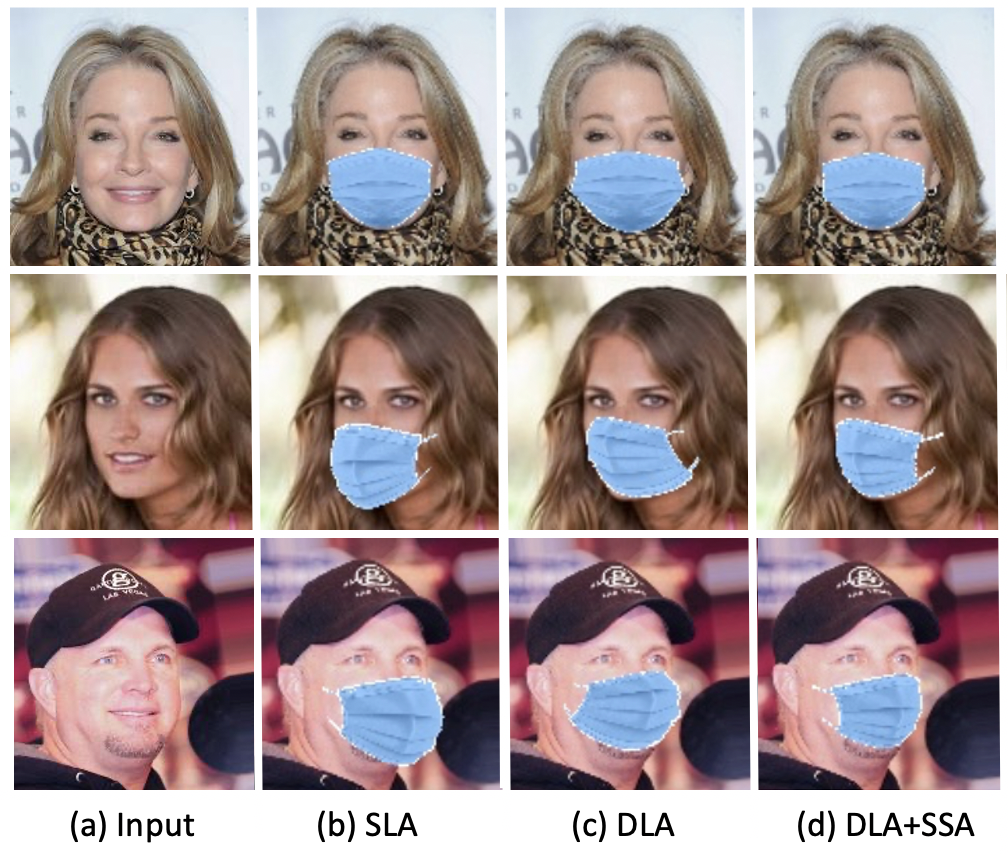}}
\caption{(a) Front and half profile face images randomly drawn from the CelebA dataset~\cite{}. Column 2-4 show mask overlay examples using (b) sparse landmark alignment (SLA) ~\cite{MaskTheFace}, (c) dense landmark alignment (DLA), and (d) the proposed method incorporating DLA and statistical shape analysis. }
\label{fig1}
\end{figure}

Prior efforts in literature usually focus on tasks related to face mask detection and recognition. There are many studies and algorithms developed to detect if people are wearing masks. A few investigate algorithms for mask removal and face inpainting. However, no further instructions are suggested for people who are wearing masks properly. This motivates us to design an automated system for face-covering monitoring and demonstration. We argue that directly demonstrating correct face-covering through visual displays, rather than text explanations, facilitates guiding the habits of wearing masks to prevent the spread of the virus, eventually beneficial to the public. For example, the proposed system could be set up at the doors of densely populated places such as airports, subway stations, or large shopping centers. If an individual is wearing a mask improperly, the person can take action using the visual demo as a reference.

The proposed system consists of two modules: face mask detection and mask overlay. Specifically, an input facial image is first classified into one of the three categories: "correctly wearing", "not wearing," and "incorrectly wearing". For "correctly wearing", the mask overlay module is bypassed, and the original image is displayed as the demonstration. Otherwise, the mask overlay module edits the input image by adding a mask to the face. To this end, two challenges need to be addressed. First, the face is not a rigid object that always stays in the same shape; Second, faces in images are usually positioned differently due to camera positions and angles. To tackle these challenges, we incorporate SSA and dense landmark alignment in our mask overlay module so that variations in face shapes, orientations, and scales are considered in mask put-on. Column (d) of Figure \ref{fig1} presents several examples of the proposed mask overlay module. Compared to the prior study, MaskTheFace~\cite{MaskTheFace}, our method significantly improves the results on half profile faces. We summarize the contributions in this paper:
\begin{itemize}
    \item An automated pipeline is proposed to monitor face-covering conditions and render a face-covering demo. The generated images are very realistic on both front and profile face images.
    \item As an essential component in our pipeline, a novel and effective mask overlay procedure is developed by fusing statistical shape analysis and dense face landmark alignment.
\end{itemize}

More visually appealing results of the proposed method are presented in the experimentation section. %Our code will be publicly available upon paper acceptance.

\section{Related work} \label{sec2}
The problem in this study involves several research topics: face mask detection, mask removal, and mask overlay/inpainting. In this section, we present a brief review of the most relevant works in literature.

\textbf{Face mask detection} has been widely studied~\cite{Prototype,Real,Face,Face2,Realtime,Intelligent,RetinaMask}. Given an input image or real-time videos, the detectors can tell if the person is wearing a mask properly or not. Applications of mask detection usually run on edge devices such as mobile phones and embedded systems. A typical example is a mobile application called “CheckYourMask”~\cite{CheckYourMask}. It allows people to take selfies to check if they are wearing their masks correctly. Since deep models in the MobileNet family demonstrate a good balance between accuracy and resource consumption, they are usually taken as the backbone in prior arts. Specifically, Venkateswarlu et al.~\cite{Face2} proposed an algorithm using MobileNet with a global pooling block for face mask detection. Vinh and Anh~\cite{Realtime} proposed an algorithm using a Haar cascade classifier to detect the face and YOLOv3 to detect the mask, achieving a 90\% detection rate. Xue et al.~\cite{Intelligent} and Jiang et al.~\cite{RetinaMask} improved the RetinaFace algorithm~\cite{RetinaFace} by inferring the positions of the mask. Oumina et al.~\cite{Real} proposed a system using ensemble learning to detect if a person is wearing a mask. They extracted face features using different deep learning models (VGG19, Xception, and MobileNetV2) and achieved the detection by fusing the classification results of SVM and KNN.

\textbf{Mask removal} is essential for images that are classified as “incorrectly wearing” in the proposed method. It removes the wrong worn masks from faces meanwhile attempts to edit the images such that complete, non-occluded faces are reconstructed and displayed. Though there are many successful segmentation and image inpainting algorithms, Din et al.~\cite{Anoval} argued that most prior approaches did not fit the problem of unmasking covered face well due to the large size of masks (e.g., face masks usually cover the front beyond the face boundary below the chin). To tackle this problem, they investigated a two-stage method where the first stage detects and segmented masks with a modified version of the U-Net and the second stage deployed a GAN-based network with global and local discriminators for mask-area inpainting.

\textbf{Mask put-on} is a process to overlay a mask in a face image. There are many ways to add masks to faces, such as manually photoshop an image in~\cite{Anoval}. Among various methods, to the best of our knowledge, MaskTheFace~\cite{MaskTheFace} is the only study to achieve this goal automatically. It uses six facial landmarks, one on the nose bridge, two on the cheeks, and three along the chin line. Then a mask template is matched to the six landmarks to overlay the mask onto faces. Since only six facial landmarks are used in this method, we call the method proposed in MaskTheFace sparse landmark alignment (SLA) in this paper. Column (b) of Figure \ref{fig1} shows examples of SLA on face images. We observe that SLA fails to follow the face boundary in half-profile images. This observation motivates us to introduce dense landmark alignment (DLA) and statistic shape analysis in our mask overlay algorithm.

\captionsetup[figure]{font=small}
\begin{figure*}\smaller
\centerline{\includegraphics[width=0.9\linewidth]{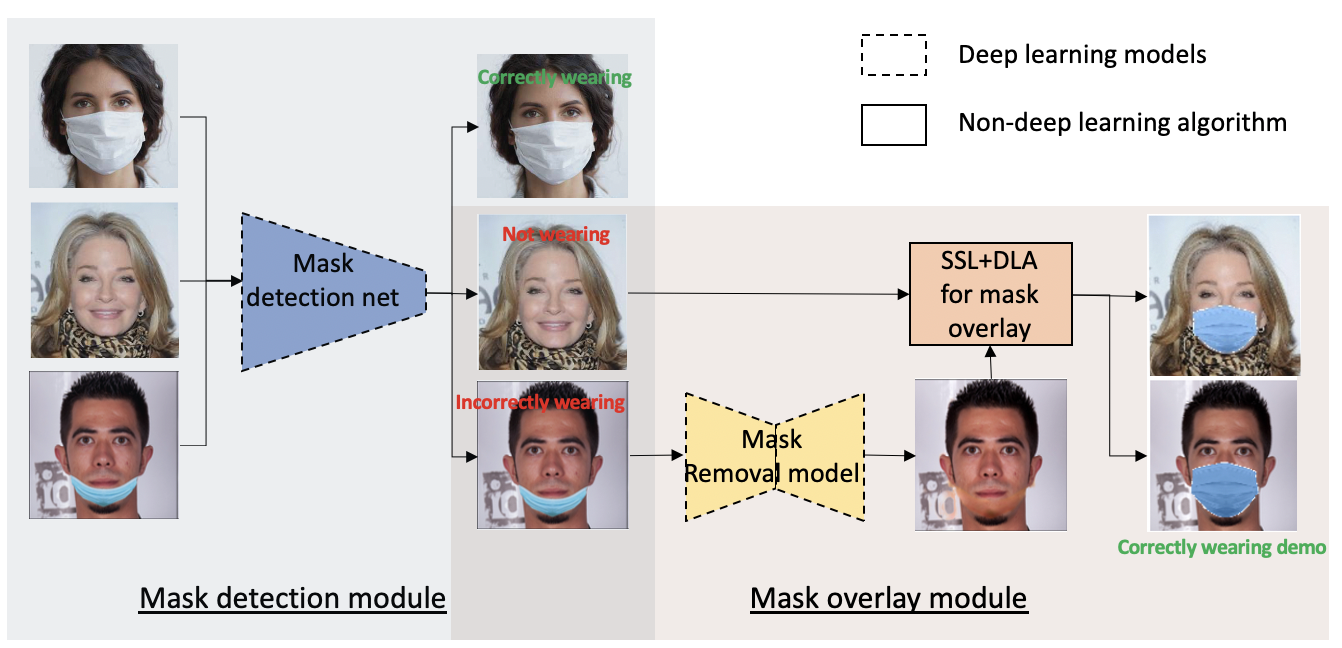}}
\caption{Systematic diagram of the proposed pipeline. For an input image, the classifier takes MobileNetV2 as the backbone for mask detection (Sec.\ref{3.1}). Based on the classification results, a mask removal model may be deployed (Sec.\ref{3.2}). The final face-covering example is generated using statistical shape analysis and dense landmark alignment (Sec.\ref{3.3}). In the diagram, we use dash-edge boxes to mark deep learning models; by contrast, the solid-edge box represents a non-deep learning algorithm.}
\label{defense}
\end{figure*}

\section{Methodology}
An overview of our pipeline is provided in Figure 2. When an image is classified as “Correctly Masked”, no further action is required. When an input is classified as “Not Masked”, a mask is overlaid on the image. Otherwise, the wrong-worn mask is removed, and a new mask is inpainted to cover the proper face region for visual demo. We will elaborate on the technical details of the two composing modules: mask detection module and mask overlay module in this section.

\subsection{Mask detection module}\label{3.1}
In our pipeline, the mask detection module classifies an input facial image into one of the three categories, “correctly wearing”, “incorrectly wearing” , and “not wearing", which are passed to the downstream mask overlay module. Following previous efforts in literature, we take MobileNetV2 as the backbone of the detector. Specifically, the MobileNetV2 model is pre-trained on ImageNet. We replace the fully connected layer at the top with a 7-by-7 average pooling layer and two dense layers with a ReLU activation function. Before the output layer, a dropout layer with a rate of 0.25 is applied in training.

\subsection{Mask overlay module}
This module comprises two algorithms: mask removal using a GAN-based model and mask put-on using statistic shape analysis and landmark alignment.

\captionsetup[figure]{font=small}
\begin{figure}\smaller
\centerline{\includegraphics[width=0.935\linewidth]{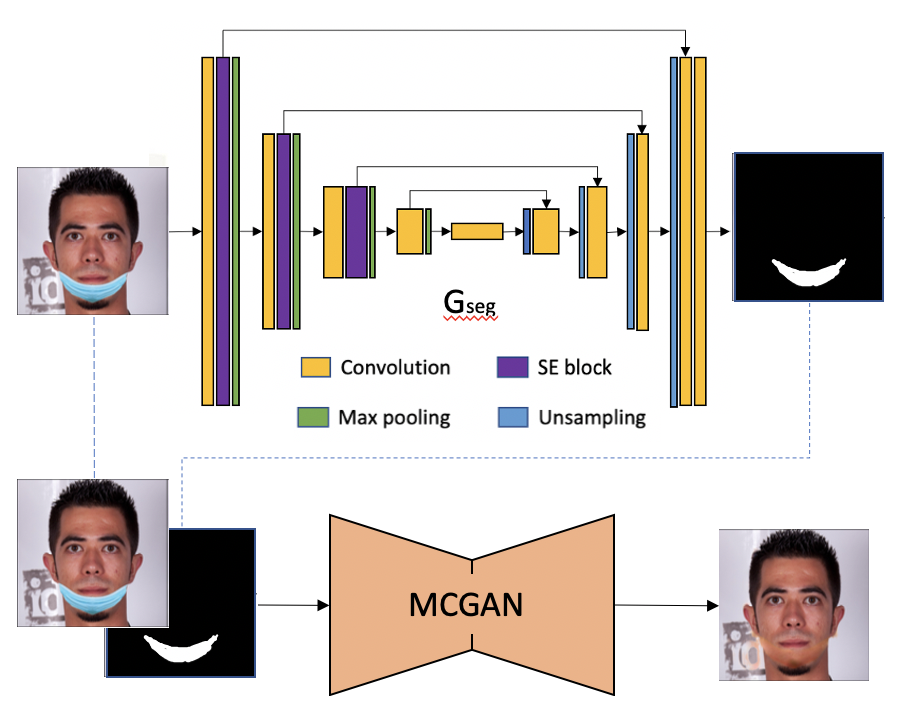}}
\caption{Our mask removal model is composed of a mask segmentation net and a face inpainting net. We modify the MCGAN structure~\cite{Interactive} for occluded face recovery.}
\label{fig3}
\end{figure}

\subsubsection{Mask removal}\label{3.2}
The purpose of the mask removal algorithm is to remove wrong-worn masks and synthesize a non-occluded face. In this regard, we adopt the MCGAN structure proposed by Khan et al.~\cite{Interactive}. Note that in MCGAN, a binary mask is required as input to recover the occluded region on the face. In this regard, we design a mask segmentation network for a binary mask. 

The specific structure of our mask removal net is depicted in figure \ref{fig3}. In the segmentation network $G_{seg}$, the squeeze-and-Excitation (SE) Block~\cite{SE} that performs re-calibration of channel characteristics is incorporated in the UNet structure. The segmentation loss combines a Dice loss, $L_{Dice}$, and a binary cross-entropy loss, $L_{BCE}$, to evaluate the similarity between the obtained binary map $I_{mask}$ and the ground-truth $I_{gt}$:
\begin{equation}
L_{seg}=L_{Dice}+L_{BCE},
\end{equation}
where $L_{Dice}$ measures the region-based similarity between $I_{mask}$ and $I_{gt}$ and $L_{BCE}$ measures the global distribution differences between the two masks.

The MCGAN model~\cite{Interactive} cascades two generative nets for face inpainting: one for maintaining face global semantics and achieving coarse face recovery, and the other for face refinement. To obtain a realistic synthesis face image, MCGAN exploits adversarial discriminators for both generative nets. The complex structure of MSGAN makes its training hard and unstable. We notice that previously covered regions in facial images will eventually be covered again in our problem. This relaxes the level of face recovery and prompts us to simplify the MCGAN structure for easy training. Specifically, we remove the adversarial discriminator paired with the coarse inpainting net. To compensate for the performance loss, we replace all convolutional layers with gated convolution blocks~\cite{GatedConv}. With a soft mask mechanism, gated convolution enables the model to learn the masked regions in a separate path, thus generating more realistic inpainting results. In addition, we deploy a pre-trained VGG-16 to quantify the perceptual loss between the inpainting face $I_{inp}$ and its ground-truth $I_{gt}$. In sum, our target function to train the modified MCGAN model is
\begin{equation}
L_{MCGAN}=L_{GAN}+\lambda_{rc}L_{rc}+\lambda_{p}L_p,
\end{equation}
where $L_{GAN}$, $L_{rc}$, and $L_{p}$ are the adversarial loss, image reconstruction loss, and perceptual loss between $I_{inp}$ and $I_{gt}$. $\lambda_{rc}$ and $\lambda_p$ are the weights of reconstruction loss and perceptual loss, respectively. Our image construction loss combines $L_{1}$ loss and image structural similarity metric (SSIM) loss:
\begin{equation}
    L_{rc}(I_{inp}, I_{gt})=|I_{inp}, I_{gt}|_1 - SSIM(I_{inp}, I_{gt}).
\end{equation}
Assuming $\phi_i$ is the activation map of the i-th layer in the pre-trained VGG-16, the perceptual loss is computed by:
\begin{equation}
L_p(I_{inp}, I_{gt})=\sum_i|\phi_i(I_{inp})-\phi_i(I_{gt})|_1.
\end{equation}

\captionsetup[figure]{font=small}
\begin{figure}\smaller
\centerline{\includegraphics[width=\linewidth]{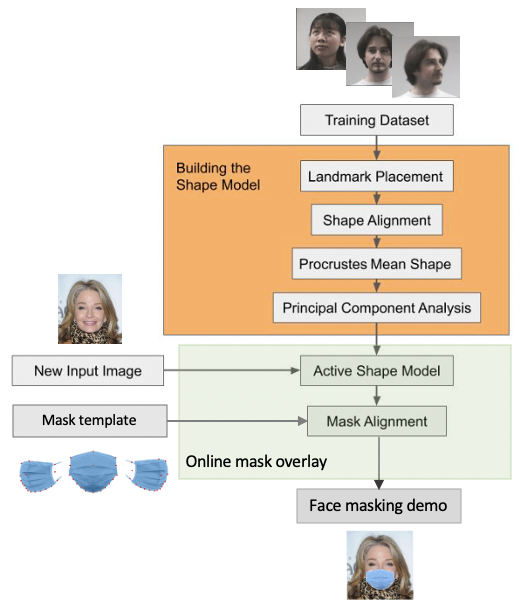}}
\caption{Block diagram of the proposed mask overlay algorithm that incorporates SSA and DLA. In the figure, the training images are from the Pointing’04 dataset [23] and test image from the CelebA image set \cite{}.}
\label{fig4}
\end{figure}

\subsubsection{Mask put-on}\label{3.3}
Mask put-on is an essential component in our pipeline. Instead of leveraging deep learning, we design a non-deep learning algorithm for mask overlay for two reasons. On the one hand, the lack of suitable training samples hinders the use of deep learning. Though both MaskedFace-Net~\cite{CheckYourMask} and MaskTheFace~\cite{MaskTheFace} provide paired non-occluded facial images and face-covering images, those face-covering images are not realistic; the noticeable distortions and artifacts in images may bias the training. On the other hand, a clear definition of proper face-covering with a mask has been elaborated in WHO guidelines: all nose, mouth, and chin should be covered. Since masks are usually in similar shape, we argue that put-on masks can be efficiently achievable by aligning face landmarks and mask templates.

To synthesize realistic face-covering images, accurate localizing landmarks in both face images and mask templates is crucial. We follow the 68 facial landmark convention in face landmark estimation and use the pre-trained facial landmark detector inside the dlib library on face images. However, we notice that landmark estimation of faces in profile is poor. Though increasing facial landmark numbers in landmark alignment improve the performance (please refer to column (c) of Figure 1 for examples of DLA), it still fails to follow the chin line in images closely. Therefore, we propose to build a face shape model using SSA, specifically active shape model (ASM)~\cite{ASM}, so that variations in face shapes, orientation, and scales due to different camera positions can be accommodated. To this end, we start by estimating landmarks on the contours of face image samples. Then we organize these face landmark coordinates in a matrix called point distribution matrix (PDM), where each row corresponds to one face. After Procrustes analysis to align the coordinates and PCA to reduce the dimension of PDM, the face ASM is formulated as:
\begin{equation}
\label{asm}
 f_i=\Bar{f}+Pb_i, 
\end{equation}
where $f_i$ and $\Bar{f}$ represent the latent landmark representations of a specific face sample and the mean face cross all training images, $P = [p_1,p_2,...,p_t]$ is the matrix consisting of the first $t$ eigenvectors in PCA, and $b = [b_1,b_2,...,b_t]$ is a vector of weights acting like “knobs” to fit a specific $f_i$. The procedure to build the shape model is briefly demonstrated in Figure \ref{fig4}. Without directly using facial landmarks in a new image $f_{new}$, we use the ASM in (\ref{asm}) to approximate the geometry of the input and estimate the corresponding landmarks by searching the best set of transition/rotation/scale parameters $(\tau,\theta,s)$:
\begin{equation}
\underset{\tau,\theta,s,b}{\mathrm{argmin}}||f_{new}-M(\Bar{f}+Pb)||^2,
\end{equation}
\[
\text{where} \hspace{0.3cm} M=s \begin{bmatrix}
\cos(\theta) & \sin(\theta)\\
-\sin(\theta) & \cos(\theta)
\end{bmatrix}+\tau
\]

\begin{figure}\smaller
\centerline{\includegraphics[width=\linewidth]{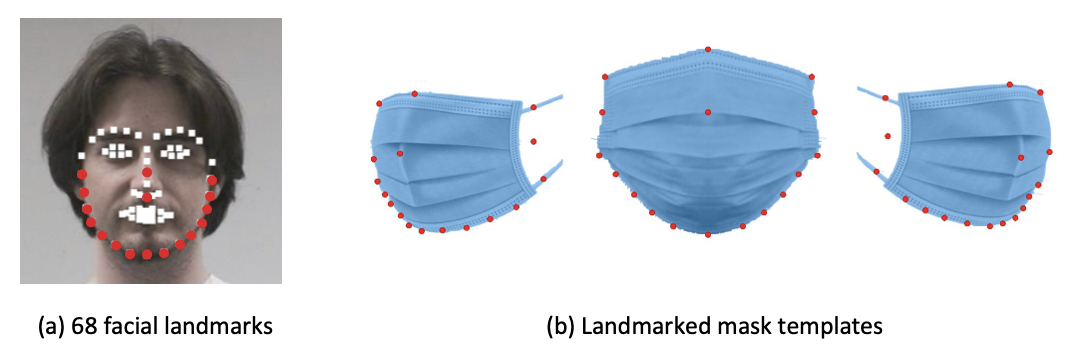}}
\caption{(a) Conventional 68 facial landmarks where the red dots are the 17 landmarks used in our mask overlay algorithm. (b) Mask templates and corresponding 17 landmarks.}
\label{fig5}
\end{figure}

As demonstrated in Figure \ref{fig4}, after obtaining the landmarks of a face in a query image, we align landmarks of the face with a mask template. In this study, we consider three mask templates, one for front view and another two for profiles. In contrast to MaskTheFace~\cite{MaskTheFace} that adopts sparse landmark alignment, we propose the use of dense landmark alignment for realistic face masking images. Because the mask should cover the bottom of the chin and the top of the nose above the nose tip, instead of using six landmarks, we manually annotate 17 landmarks on mask templates that correspond to face landmarks ranging from index 2 to 16 and 30 and 34 in the conventional 68-landmark patterns. We illustrate the traditional 68 face landmarks and our annotated mask templates in Figure \ref{fig5}.

\captionsetup[figure]{font=small}

\section{Experiments and Discussions}
Our whole pipeline consists of two modules: face mask detection and mask overlay. Since the two modules are relatively independent, we evaluate their performance separately.

\subsection{Face mask detection}
\textbf{Data set:}
We collect 5829 images from two public datasets, MaskedFace-Net~\cite{CheckYourMask} and Flickr-Faces-HQ Dataset (FFHQ)~\cite{FFHQ}. MaskedFace-Net is a large synthetic dataset consisting of paired correctly and incorrectly face-covering photos. Images in MaskedFace-Net contain considerable variation in age, ethnicity, and image background. We randomly pick 1903 images with correct mask-wearing and 1926 images of wrong mask-wearing from the dataset; paired images are avoided. In addition, 2000 non-occluded images are randomly selected from the FFHQ image set.

In our experiment, all 5829 images are resized to 224-by-224. Images in each category are randomly divided into a training and testing set with a ratio of 8:2, resulting in 4663 training samples and 1166 testing images.

\textbf{Implementation details:} Category cross-entropy is used to train our mask detector. We use Adam with a learning rate of $10^{-4}$ and a decay rate of $5\times10^{-6}$ to optimize the model. The model is set to be trained maximum for 20 epochs with a batch size of 32. The early stop mechanism is deployed to prevent overfitting. 

\textbf{Results:} Our face-covering monitoring model achieves a 98\% detection rate. The specific results in each category are summarized in Table \ref{T1}.

\begin{table}[htbp]\smaller
\caption{Performance of our triple-category facial mask detection.}
\vspace{-0.3cm}
\begin{center}
\begin{tabular}{ c|c|c|c|c } 
 \hline
   & Image \# & Precision & Recall & F-1 score \\
 \hline
 Correctly wearing & 381 & 0.98 & 0.96 & 0.97 \\  \hline
 Incorrectly wearing & 385&  0.96 & 0.98 & 0.97 \\  \hline
Not wearing & 400&1.00  & 1.00 & 1.00 \\
 \hline
\end{tabular}
\label{T1}
\end{center}
\end{table}

\subsection{Mask overlay}
\textbf{Data sets:}
Our mask overlay module is applied to images classified as either "not wearing" or "incorrectly wearing". To train the mask removal model, we randomly pick ten thousand images from the public CelebA dataset~\cite{CelebA}. All of these images are resized to 512-by-512, with the faces centered in images. We manually add masks on images to obtain paired samples of non-occluded faces, masked faces and mask binary maps. The dataset is partitioned with an 8:2 ratio for training/validation.

To build our face shape model, we select a subset of 1950 data samples (130 examples each of 15 individuals) from the head pose estimation dataset Pointing’04~\cite{Pointing04}. These photos are taken by changing the orientation of the head in the direction vertically and horizontally.

To evaluate the generalization of our mask overlay module, the test images of "no mask" and "incorrectly mask wearing" are drawn from celecA~\cite{CelebA} and MaskedFace-Net~\cite{CheckYourMask} during test, respectively.

\textbf{Implementation details:} To train our mask removal model, we set the batch size of 4 and 2 for mask segmentation and face inpainting, respectively. The optimizer for both models is Adam, with a learning rate of 0.001. The segmentation model is trained for 200,000 iterations, and the inpainting model is trained for 300,000 iterations. Since our mask-overlay algorithm is not based on deep learning, we directly feed our Pointing'04 subset into ASM. Then images with non-occluded faces are passed to the mask overlay algorithm for image editing.

\captionsetup[figure]{font=small}
\begin{figure}\smaller
\centerline{\includegraphics[width=0.9\linewidth]{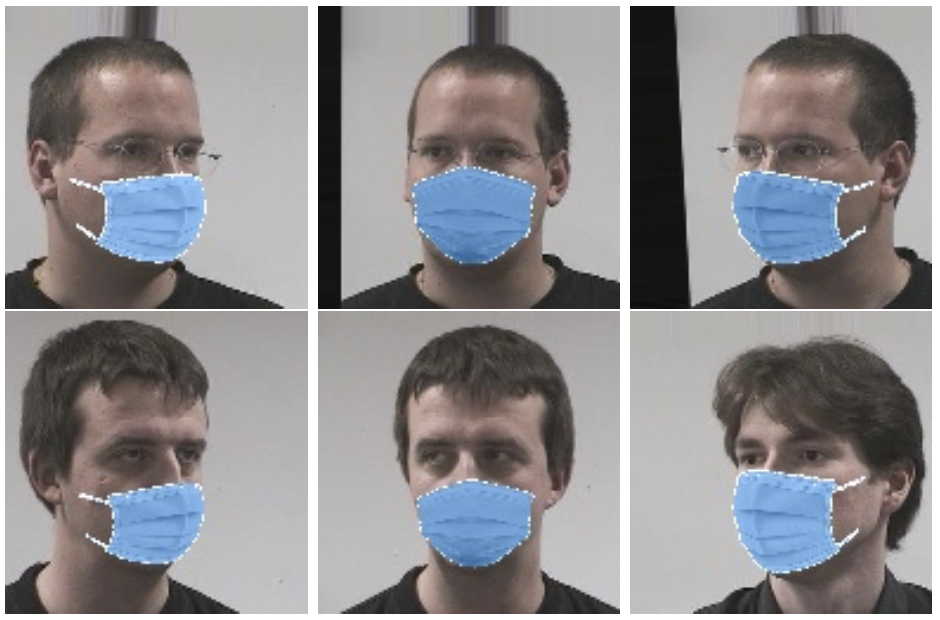}}
\caption{Examples of our mask overlay results on non-occluded face images in different head orientations. The face images are randomly picked from the Pointing’04 dataset~\cite{Pointing04}.}
\label{fig6}
\end{figure}

\begin{figure}\smaller
\centerline{\includegraphics[width=0.9\linewidth]{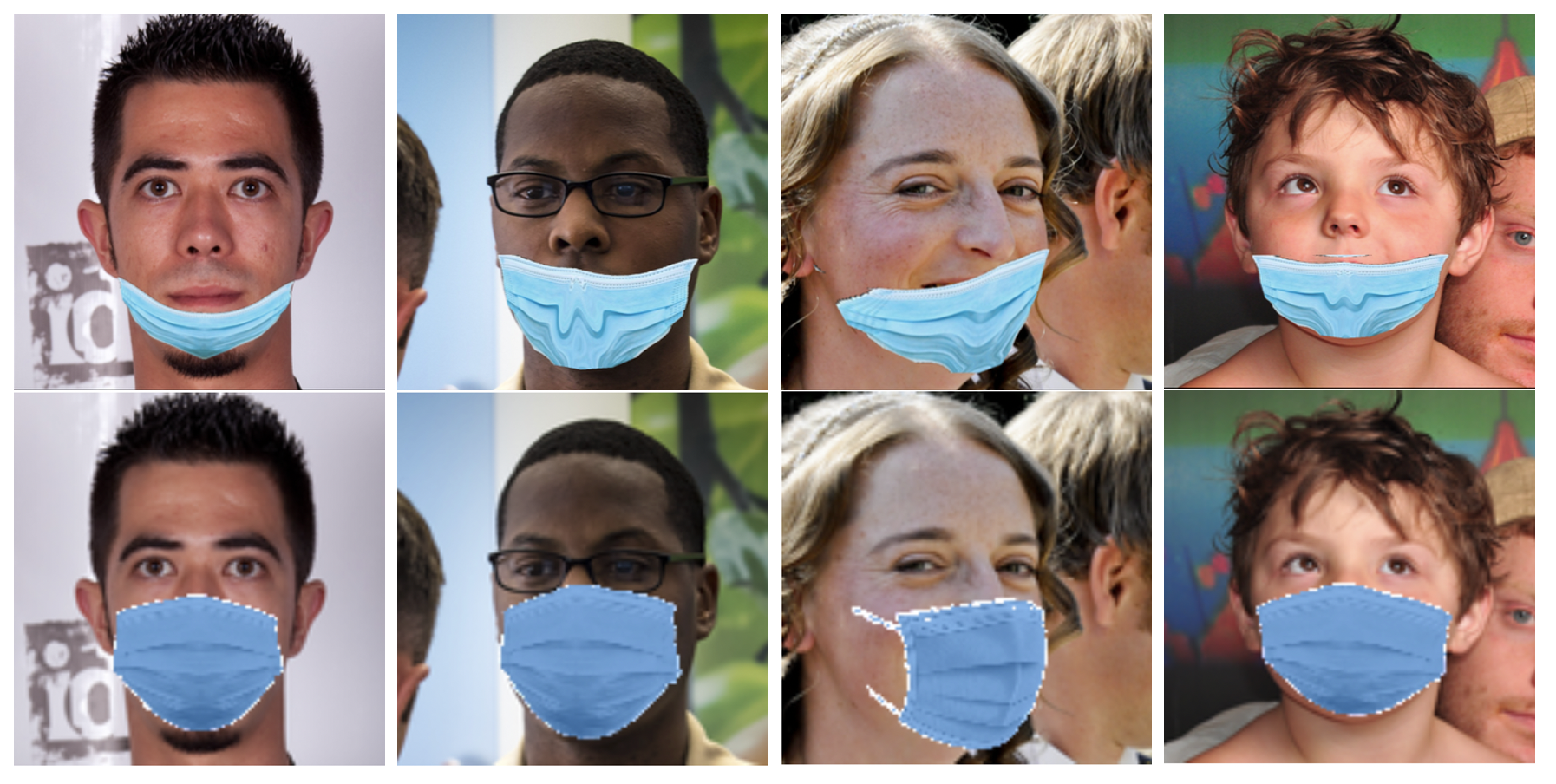}}
\caption{First row: images with "incorrectly mask-wearing" from the MaskedFace-Net~\cite{CheckYourMask}. Second row: mask overlay results generated by our method.}
\label{fig7}
\end{figure}

\textbf{Results:} Figure \ref{fig6} presents examples of our mask overlay results. In literature, MaskTheFace~/cite{MaskTheFace} is the only study to achieve the goal of mask overlap on non-occluded images automatically. We compare the SLA algorithm in MaskTheFace and the proposed method and present the results in column (b) and column (d) in Figure \ref{fig1}. Visually, our mask overlay algorithm that incorporates SSA and DLA obtain noticeable improvement on faces with different orientations. Figure \ref{fig7} presents results of our mask overlay on images with incorrect mask-wearing.

\captionsetup[figure]{font=small}

\subsection{Ablation on mask overlay}
In this experiment, we take the SLA algorithm in MaskTheFace~\cite{MaskTheFace} as the baseline and investigate the effect of DLA and SSM on the final mask overlay results. To this end, we take non-occluded facial images from the CelecA image set~\cite{CelecA} and generate face-covering images using the SLA algorithm, DLA algorithm based on the 17 landmarks without the face shape model, and our DLA+SSA approach. Examples of mask overlay are displayed in Figure \ref{fig1}. Comparing to sparse landmark alignment, dense landmark alignment helps to fit the mask to faces. The active shape model with dense landmark alignment generates more realistic face-covering images.

\section{Conclusion}
We considered an important problem of monitoring and demonstrating face-covering conditions from images. Our approach was capable of detecting face images with improperly mask-wearing and rendering plausible personalized face-covering demonstrations. Experimentation showed that the proposed mask overlay algorithm based on active shape model and dense landmark alignment outperformed prior arts. For future work, we plan to test our pipeline on diverse images taken under different illumination (i.e. shadows, reflections, etc.). 

%\newpage
{\small
%\bibliographystyle{ieee_fullname}
%\bibliography{egbib}

}

%
%{\small
%\bibliographystyle{ieee_fullname}
%\bibliography{egbib}
%}

\end{document}